%% file: main.tex
\documentclass[sigconf,nonacm]{acmart}

\usepackage{booktabs} 
\usepackage{graphicx}

\usepackage{cuted}
\usepackage{capt-of}
\usepackage{multirow}
\usepackage[ruled]{algorithm2e} 

\SetAlFnt{\small}
\SetAlCapFnt{\small}
\SetAlCapNameFnt{\small}
\SetAlCapHSkip{0pt}

\def\papername{X-Actor}

\begin{document}
\title{\papername: Emotional and Expressive Long-Range Portrait Acting from Audio}

\author{
    Chenxu Zhang,\quad
    Zenan Li,\quad
    Hongyi Xu,\quad
    You Xie,\quad
    Xiaochen Zhao,\quad
    Tianpei Gu,\quad\quad\quad
    Guoxian Song,\quad
    Xin Chen,\quad
    Chao Liang,\quad
    Jianwen Jiang,\quad
    Linjie Luo\quad
}
\affiliation{%
 \institution{Bytedance Intelligent Creation}
 \country{USA}
}
\email{{chenxuzhang; zenan.li; hongyixu; you.xie; xiaochen.zhao; tianpei.gu; guoxiansong; chris.chen1; linjie.luo}@bytedance.com}

\renewcommand\shortauthors{Chenxu Zhang et al.}

\begin{strip}
	\centering
	\includegraphics[width=0.97\linewidth]{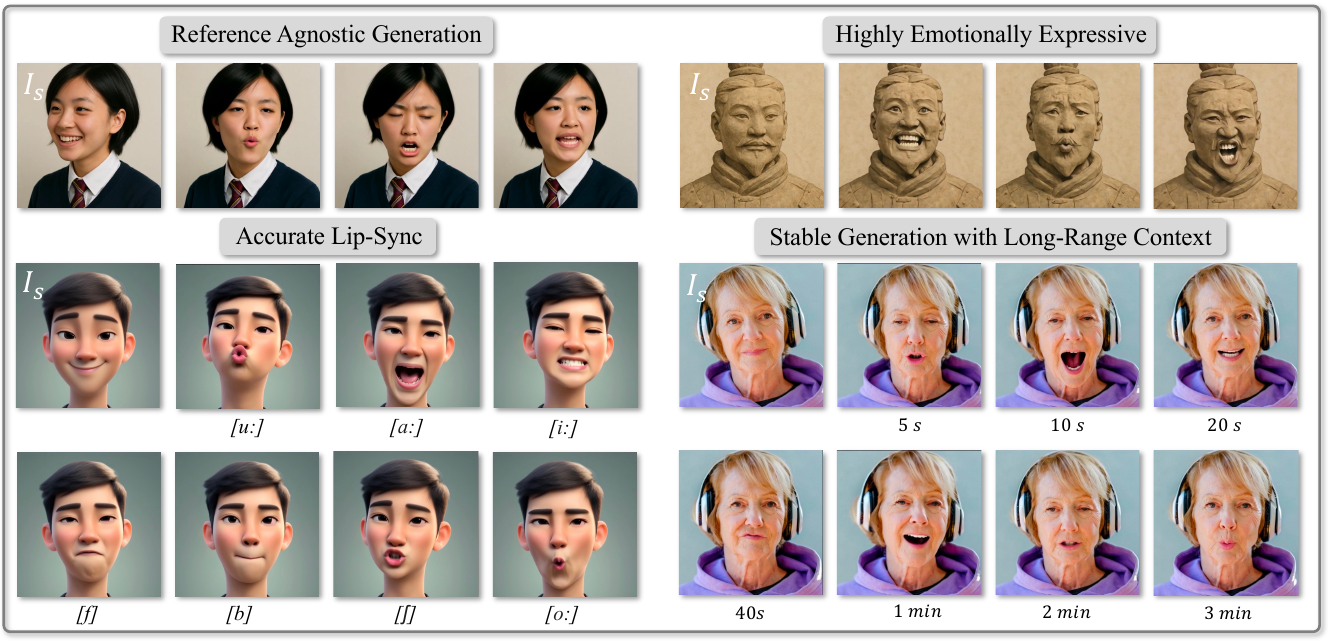}
	\centering
	\captionof{figure}{We present \papername, an audio-conditioned autoregressive diffusion framework for generating emotionally expressive, highly dynamic and lifelike portrait performances, while maintaining accurate long-form lip synchronization and consistent emotion-audio alignment.} 
    \label{fig:teaser}
\end{strip}

\begin{abstract}
We present \papername, a novel audio-driven portrait animation framework that generates lifelike, emotionally expressive talking head videos from a single reference image and an input audio clip. 
Unlike prior methods that emphasize lip synchronization and short-range visual fidelity in constrained speaking scenarios, \papername~enables actor-quality, long-form portrait performance—capturing nuanced, dynamically evolving emotions that flow coherently with the rhythm and content of speech.
Central to our approach is a two-stage decoupled generation pipeline: an audio-conditioned autoregressive diffusion model that predicts expressive yet identity-agnostic facial motion latent tokens within a long temporal context window, followed by a diffusion-based video synthesis module that translates these motions into high-fidelity video animations. By operating in a compact facial motion latent space decoupled from visual and identity cues, our autoregressive diffusion model effectively captures long-range correlations between audio and facial dynamics through a diffusion-forcing training paradigm, enabling infinite-length emotionally-rich motion prediction without error accumulation. Extensive experiments demonstrate that \papername~produces compelling, cinematic-style performances that go beyond standard talking head animations and achieves state-of-the-art results in long-range, audio-driven emotional portrait acting. Please refer to \protect{\url{https://byteaigc.github.io/X-Actor/}} for more results.
\end{abstract}

\maketitle

\input{sections/introduction}
\input{sections/relatedworks}
\input{sections/method}
\input{sections/experiments}
\input{sections/conclusion}

\bibliographystyle{ACM-Reference-Format}
\bibliography{main}

\clearpage

\begin{figure*}[t]
\centering
\includegraphics[width=0.95\linewidth]{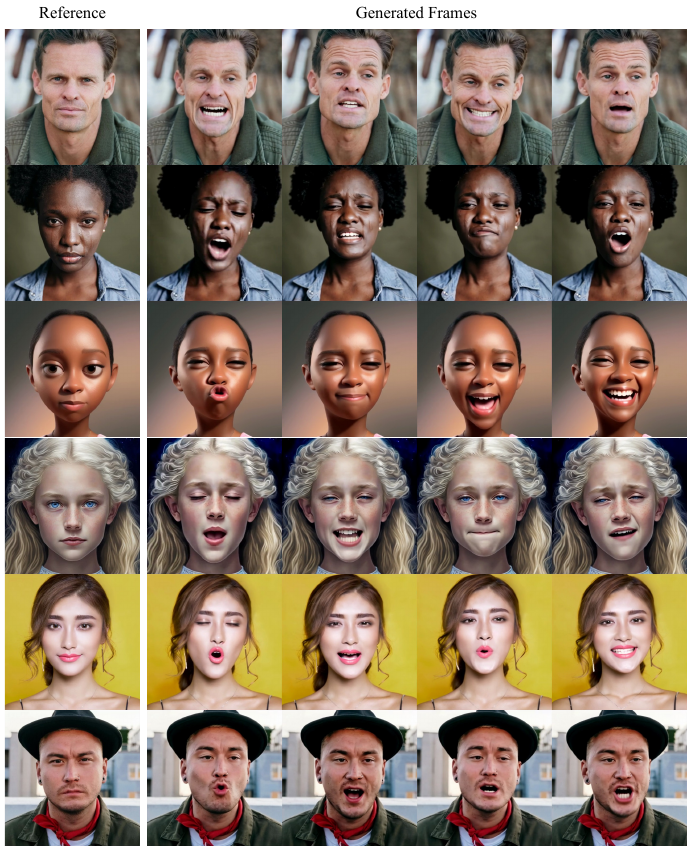}
\vspace{-4mm}
\caption{More qualitative results. The leftmost column shows the reference image, and we present the talking face animations driven by audio inputs.}
\label{fig:big1}
\end{figure*}

\begin{figure*}[t]
\centering
\includegraphics[width=0.95\linewidth]{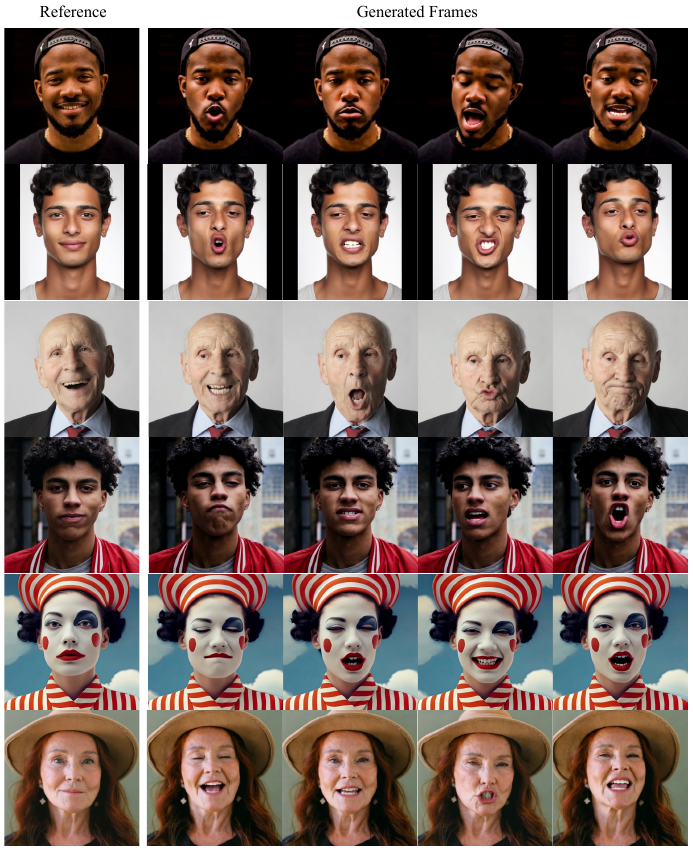}
\vspace{-4mm}
\caption{Additional qualitative results. The leftmost column shows the reference image, and we present the talking face animations driven by audio inputs.}
\label{fig:big2}
\end{figure*}
\end{document}

%% file: sections/introduction.tex
\section{Introduction}
Animating human portraits from audio is a long-standing challenge at the intersection of computer vision, graphics, and multimodal learning. Recent advances in talking head generation have made significant strides in producing visually realistic facial animations synchronized with speech~\cite{xu2024vasa,tian2024emo,chen2025echomimic,hallo,lin2025omnihuman,ji2024sonic}. However, most existing methods are limited in scope—focusing primarily on lip synchronization and short-range facial dynamics in constrained talking scenarios. These approaches fall short in delivering actor-level performances that involve emotional depth, long-range temporal coherence, and context-aware motion evolution. Such capabilities are critical for applications ranging from virtual avatars and digital actors to film dubbing and immersive telepresence.

In this work, we introduce \papername, a novel framework for long-range emotional and expressive portrait acting from audio. Unlike conventional talking head models that prioritize short-term lip sync and visual fidelity, \papername~aims to bridge the gap between speech-driven animation and actor-quality emotional performance. Our goal is to generate full-spectrum facial behavior—encompassing subtle expressions, emotional intensity, natural head dynamics, and fluid emotion transitions—that evolve over long time horizons and align seamlessly with the rhythm and semantics of the input audio.

To this end, we propose a two-stage decoupled generation pipeline comprising motion acting and visual synthesis. 
The first stage features an audio-conditioned, chunk-wise autoregressive diffusion model that generates expressive facial motion in a continuous latent space, decoupled from visual appearance and identity. By denoising motion latents chunk by chunk while attending to long-range audio and motion context, the model captures temporally coherent, audio-aligned, emotionally-rich facial dynamics over extended sequences.
In the second stage, these motion latents, combined with a single reference portrait, are transformed into high-fidelity, identity-preserving video frames using a pretrained diffusion-based video generator. 
This modular architecture separates the challenges of long-context audio-motion alignment from those of photorealistic rendering, leveraging autoregressive models for long-range, cross-modal dynamics, and diffusion models for visual fidelity.
Crucially, our facial motion latents are compact and semantically structured to capture nuanced dynamics while abstracting away appearance and identity. Unlike approaches that operate in the pixel latent space, our autoregressive model focuses exclusively on learning the long-form dynamic correlation between facial expressions and audio flow from large-scale, variable-length talking sequences from diverse speakers. This design also ensures that motion generation is agnostic to the reference portrait’s identity or expression, enabling adaptive responses to a wide spectrum of emotional audio input—even when the reference portrait displays conflicting emotions.

A key challenge in long-context, audio-driven portrait animation—whether using diffusion with sliding-window context or autoregressive prediction—is the non-recoverable quality degradation and context drift over time, primarily due to accumulated generation errors.  
To address this, we integrate a diffusion-force paradigm~\cite{chen2024diffusion} to our autoregressive diffusion. Specifically, each motion latent chunk is trained independently and in parallel to denoise using a chunk-specific timestep while conditioning on a noised historical motion context. This strategy effectively mitigates inference-time drift, reducing the reliance on clean ground-truth history, and enables stable, high-quality motion generation over arbitrarily long durations. Additionally, we incorporate inference-time strategies such as time-adaptive history context guidance schedulers and Classifier-Free Guidance (CFG) to further promote expressive, emotion-aware portrait acting.

Trained on a large and diverse corpus of talking head videos with varying durations, \papername~not only achieves accurate and consistent lip synchronization with input audio, but also significantly outperforms existing methods in capturing emotional expressiveness, long-range facial dynamics, and temporal coherence. Qualitative results further demonstrate that our model produces cinematic, actor-quality performances, effectively capturing fine-grained shifts in speech and sentiment. 
In summary, our main contributions are:
\begin{itemize}
\setlength{\itemsep}{3pt}
\setlength{\parskip}{0pt}
\setlength{\parsep}{0pt}
\item A novel framework for generating long-range, emotionally expressive portrait animations from audio, producing actor-quality talking head videos from a single reference image.
\item A modular architecture that decouples motion generation from video synthesis, enabling reference-agnostic and emotion-aware audio control.
\item Integration of a diffusion-forcing training paradigm into autoregressive diffusion over a compact, continuous motion latent space, enabling stable, infinite-length motion generation without error accumulation.
\item State-of-the-art both visual and numerical performance on long-form, emotionally rich speech datasets.
\end{itemize}

\begin{figure*}[ht]
    \centering
    \includegraphics[width=0.95\textwidth]{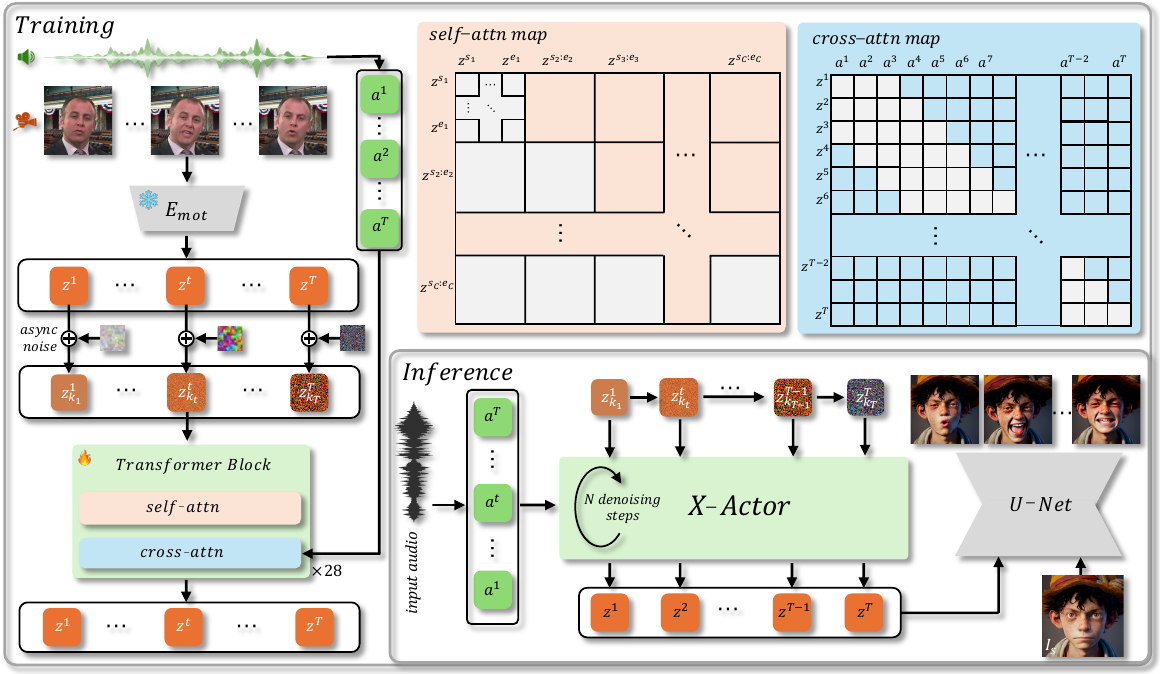} 
    \caption{\textbf{Overview of~\papername.} 
    \papername~decouples video synthesis from audio-conditioned motion generation, operating in a compact, expressive, and identity-agnostic facial motion latent space. Specifically, we encode talking video frames into sequences of motion latents $\mathbf{z}^{1:T}$ using a pretrained motion encoder $E_{\text{mot}}$. These latents are corrupted with asynchronously sampled noise levels and denoised using an autoregressive diffusion model trained with a diffusion-forcing scheme~\cite{chen2024diffusion}. Within each motion chunk $\mathbf{z}^{s_c:e_c}$, we apply full self-attention to preserve fine-grained expressiveness, while causal cross-chunk attention ensures long-range temporal coherence and context awareness. Each motion token $z^t$ attends to frame-aligned audio features $\mathbf{a}^{t-2:t+2}$ via windowed cross-attention, enabling accurate lip synchronization and capturing transient emotional shifts. At inference time, we autoregressively and iteratively predict future motion tokens with a monotonically decreasing noise schedule over the historical motion context. Finally, alongside a single reference image, we render the predicted motion sequence into high-fidelity, emotionally rich video frames using a pretrained diffusion-based video generator.}
    \label{fig:overview}
\end{figure*}

%% file: sections/relatedworks.tex
\section{Related Work}
\noindent\textbf{Audio-Conditioned Facial Motion Generation.}
Early approaches focused on direct audio-to-lip mapping while keeping other facial attributes static~\cite{chen2018lip,cheng2022videoretalking,prajwal2020lip,suwajanakorn2017synthesizing,yin2022styleheat}. These methods, however, fail to capture holistic facial behavior such as natural head motion and expressive facial dynamics. To address this limitation, later methods introduced intermediate motion representations—such as 2D landmarks, 3D blendshapes, or 3D morphable models—and decomposed the task into two stages: audio-to-motion and motion-to-video~\cite{chen2025echomimic,sun2023vividtalk,wang2024v,wei2024aniportrait,zhang2023sadtalker,zhou2020makelttalk,zhang2021facial}. While more structured, these explicit representations often lack the expressiveness needed to convey subtle emotions and nuanced micro-expressions.
A more recent line of work~\cite{xu2024vasa,he2023gaia,ma2023dreamtalk,zhang2023dream,cui2024hallo2,zhen2025teller} uses implicit latent motion representations learned from self-supervised video reenactment tasks~\cite{wang2021facevid2vid,drobyshev2022megaportraits,xnemo2025}. These representations are typically generated via audio-conditioned diffusion models and translated into video using GAN-based renderers. Although this improves realism over rigid representations, these approaches still struggle to capture fine-grained dynamics and emotional richness, and are further constrained by the limited visual fidelity of GANs.
In contrast, our method adopts a highly expressive, identity-disentangled, and compact latent motion representation, inspired by the end-to-end diffusion-based learning paradigm of X-NeMo~\cite{xnemo2025}. We combine this with a powerful pretrained diffusion model for video synthesis~\cite{rombach2022high,guo2023animatediff}, and crucially, introduce an autoregressive diffusion framework to model long-range audio-motion dependencies while maintaining fine-grained local expressiveness.

\noindent\textbf{End-to-End Diffusion Model.}
Recent methods such as EMO~\cite{tian2024emo,tian2025emo2}, Hallo~\cite{hallo,cui2024hallo3}, Loopy~\cite{jiang2024loopy}, FantasyTalking~\cite{wang2025fantasytalking}, and OmniHuman~\cite{lin2025omnihuman} leverage end-to-end diffusion models conditioned on audio to synthesize portrait videos. Most of these models autoregressively generate motion using short-range sliding motion frames~\cite{tian2024emo,hallo,lin2025omnihuman}, limiting their ability to capture long-range temporal dependencies or evolving emotional dynamics. Loopy~\cite{jiang2024loopy} extends the receptive field using temporal segments to model inter-clip relationships. More recently, MEMO~\cite{zheng2024memo} incorporates a memory-guided temporal module, and Sonic~\cite{ji2024sonic} enhances long-range modeling with global audio context.
Our approach differs fundamentally from these end-to-end frameworks by decoupling motion generation and video synthesis. This decoupling allows us to avoid entangling the reference image’s expressions with generated motion, ensuring that facial behavior is determined exclusively by audio cues. Moreover, our autoregressive diffusion model enables efficient and effective modeling of long-form audio-motion alignment, facilitating smooth and emotionally coherent animations over extended sequences.

%% file: sections/method.tex
\section{Method}

Given a single portrait image $I_{s}$ and an input audio clip ranging from a few seconds to several minutes, our goal is to generate a cinematic, temporally coherent video in which the portrait performs in sync with the audio’s content, rhythm, and emotional dynamics. To this end, we formulate the task as a two-stage decoupled generation pipeline (Section~\ref{sec:pipe}) comprising autoregressive motion generation followed by motion-conditioned video synthesis, as illustrated in our overview figure (Figure~\ref{fig:overview}). While our video diffusion stage builds on prior architectures and training strategies (Section~\ref{sec:pre}), our key contribution lies in the design of an audio-conditioned autoregressive diffusion model that operates in a compact and expressive latent space of identity-agnostic facial motion (Section~\ref{sec:ar}). This enables the generation of long-range, emotion-rich, and audio-synchronized motion sequences with coherent temporal context. To mitigate error accumulation during long-range prediction, we introduce a diffusion-forcing training scheme and optimize inference strategies tailored to our task (Section~\ref{sec:df}).

\subsection{Decoupled Cross-Conditional Generation Pipeline}
\label{sec:pipe}
Prior audio-driven video synthesis works~\cite{jiang2024loopy, hallo,wang2025fantasytalking,ji2024sonic,lin2025omnihuman} typically model the distribution of pixel-level video latents conditioned on audio features using cross-modal diffusion frameworks. While these methods achieve strong video quality and accurate lip-audio synchronization, they fall short in several key aspects. Audio primarily correlates with structural facial motion—such as lip movement, expressions, and head poses—rather than high-frequency visual details like lighting variations or hair dynamics. Modeling dense audio-to-pixel relationships across diverse speaker videos is data-inefficient and hinders generalization. Moreover, joint modeling of motion and appearance tends to entangle expressions and emotions from the reference image $I_s$, since $I_s$ and the audio $A$ are sourced from the same clip during self-supervised training. This coupling can cause inconsistencies at inference time, particularly when the emotional tone of the reference image diverges from that of the input audio. Lastly, representing each frame in the visual latent space requires hundreds of tokens, making long-range modeling computationally prohibitive and limiting emotional coherence and natural behavioral transitions in long-form speech.

To address these challenges, we decouple audio-conditioned facial motion generation from video synthesis: the diffusion model handles detailed appearance dynamics (Section~\ref{sec:pre}), while motion is predicted externally. Crucially, we seek motion representations that retain rich motion cues from unlabeled talking videos—capturing both emotionally intense expressions (e.g., frowning, staring) and subtle micro-expressions (e.g., sighing, lip pursing), as well as head poses. A compact and expressive motion representation enables efficient long-range modeling under tight computational budgets. Moreover, disentangling identity information from motion latents improves generalization across diverse speakers, allowing the model to learn universal audio-to-expression correlations.

To this end, we follow X-NeMo~\cite{xnemo2025} in training an expressive, identity-agnostic facial motion representation, where an image encoder $E_{\text{mot}}$ maps multiscale facial motions into a compact 1D latent code $\mathbf{z} \in \mathbb{R}^{512}$. The motion encoder is trained end-to-end, jointly optimized with a diffusion UNet conditioned on $\mathbf{z}$ and a reference image $I_s$ (Section~\ref{sec:pre}). To improve generalization in our motion prediction setting, we further introduce a KL divergence loss on $\mathbf{z}$ to regularize the latent space.

\vspace{-2mm}
\subsubsection{Latent Motion Conditioned Portrait Video Synthesis}
\label{sec:pre}
Our video synthesis module is built on the Latent Diffusion Model (LDM) paradigm, where a pretrained autoencoder encodes images into a compact latent space, allowing efficient generation by denoising Gaussian noise through a diffusion backbone. We adopt the Stable Diffusion UNet architecture~\cite{rombach2022high} as the generative core, enhanced with intervened spatial transformer blocks and temporal attention layers to ensure temporal consistency, following AnimateDiff~\cite{guo2023animatediff}.
To preserve identity and appearance, we follow standard practices in portrait animation~\cite{cao2023masactrl,hu2024animate}, incorporating a reference network that mirrors the UNet architecture to extract multi-scale features from the input reference image $I_s$. These features are cross-attended within the UNet’s self-attention layers. 
The predicted audio-conditioned motion latents $\mathbf{z}$ are then used to guide video generation through newly introduced cross-attention layers.
Following~\cite{xnemo2025}, the full model, including the facial motion encoder $E_{\text{mot}}$, the UNet with temporal layers, and the motion-latent cross-attention modules, is trained end-to-end using a standard diffusion loss on large-scale facial video datasets that feature diverse expressions, emotions, and head poses.

\subsection{Autoregressive Latent Motion Diffusion}
\label{sec:ar}
To model long-form evolving facial motion, we represent each training video as a sequence of facial motion latents ${\mathbf{z}^t}$, where each $\mathbf{z}^t$ is encoded by a pretrained motion encoder $E{\text{mot}}$, and $T$ denotes the total number of frames. The accompanying audio is embedded using Wav2Vec2~\cite{baevski2020wav2vec}, and the resulting intermediate and output features are fused and projected through a MLP to produce frame-aligned audio embeddings ${\mathbf{a}^{1:T}}$.
To this end, we design a temporal autoregressive model that estimates a multinomial distribution over possible future motion latents, conditioned on both the preceding motion sequence and the aligned audio embeddings.
Specifically, we adopt the Qwen‑2.5 LLM backbone~\cite{qwen2.5} and train it to estimate the joint distribution of ${\mathbf{z}^{1:T}}$,
\begin{equation}
    \phi(\mathbf{z}^{1:T} | \mathbf{a}^{1:T}) = \prod_{c=1}^{C} \phi(\mathbf{z}^{s_c:e_c} | \mathbf{z}^{<s_c}, \mathbf{a}^{<e_c}),
\end{equation}
where the sequence is divided into $C$ chunks, and $s_c$, $e_c$ denote the start and end frame indices of the $c$-th chunk. 

\begin{figure*}[ht]
    \centering
    \includegraphics[width=\textwidth]{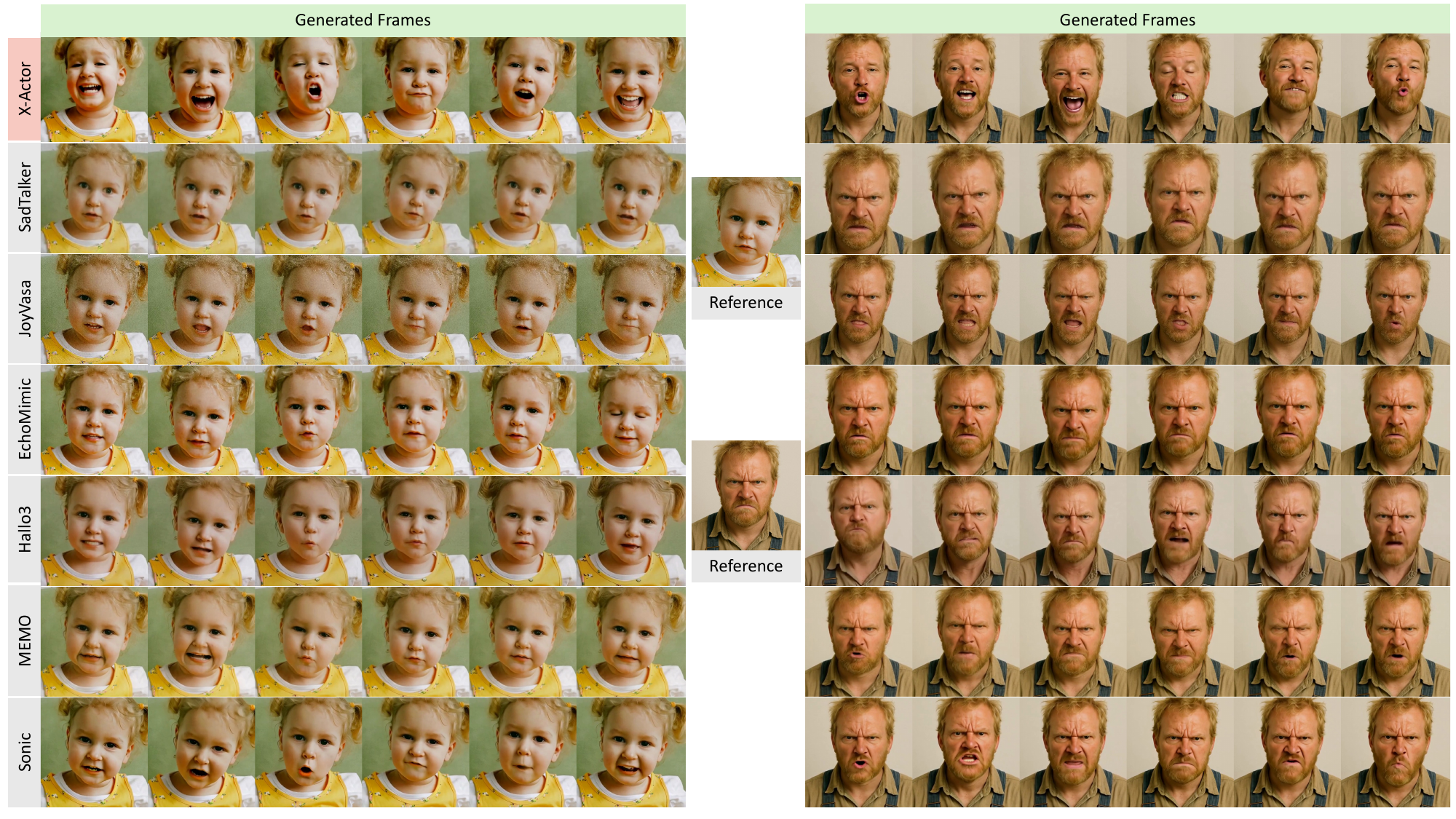} 
    \vspace{-5mm}
    \caption{Qualitative comparisons of \papername~ and SOTA audio-driven portrait animation baselines. The left example is driven by a sad and angry audio clip, while the right features a motivational presentation. Our method produces emotionally expressive performances that align well with the input audio, whereas the baselines struggle to depict consistent facial emotions and expressions, with their outputs adversely influenced by the emotion present in the reference image. }
    \vspace{-3mm}
    \label{fig:comp}
\end{figure*}

We structure the generation as a chunk-wise autoregressive diffusion process. Within each chunk, full attention is applied across all motion latent tokens to ensure consistency and stability, while temporal causal attention connects to prior generated motion chunks for long-range context modeling.
To incorporate audio conditioning, we insert dedicated cross-attention layers into the Qwen backbone, enabling each motion token $\mathbf{z}^t$ to attend to a localized, frame-aligned window of  audio embeddings $\mathbf{a}^{t-2:t+2}$, facilitating the learning of precise audio-lip synchronization. Notably, during the prediction of a motion chunk, the model also retains indirect access to broader historical audio context via inter-chunk causal self-attention, and to the full audio content of the current chunk via intra-chunk full attention, enabling audio-synchronized and contextually coherent motion generation. This stands in contrast to prior audio-conditioned diffusion models~\cite{tian2024emo,hallo,lin2025omnihuman} that rely on short sliding-window strategies (typically ~0.2s) for enforcing chunk-wise smoothness. Such “short-window” conditioning limits context to immediate audio, leading to flat, monotonic performances with little expressive buildup. In contrast, our framework leverages long-range histories—on the order of tens of seconds—allowing past audio events to inform present behavior and enabling emotionally dynamic, temporally coherent motions.

Unlike standard autoregressive models that operate in a discrete token space and are trained using cross-entropy loss, we adopt a diffusion-based training objective tailored for continuous latent motion prediction. Specifically, each motion latent chunk $\mathbf{z}^{s_c:e_c}$ is corrupted via $k$ steps of a forward diffusion process to produce a noisy sample $\mathbf{z}^{s_c:e_c}_k$. The model is then trained to predict the velocity noise component $\mathbf{v}_k$ using the v-prediction objective:
\begin{equation}
\label{eq:tf}
\mathbb{E}_{\mathbf{z}^{s_c:e_c}, \epsilon_k, k} \left[\left\|\,\mathbf{v}_k - v_\theta\left(\mathbf{z}_{k}^{s_c:e_c}, k,  \mathbf{z}^{<s_c}, \mathbf{a}^{<e_c}\right)\,\right\|^2\right],
\end{equation}
where $v_\theta$ denotes the model’s estimate of the velocity noise at step $k$, conditioned on the current noisy latents $\mathbf{z}_{k}^{s_c:e_c}$, the diffusion timestep $k$, prior motion $\mathbf{z}^{<s_c}$, and preceding audio context $\mathbf{a}^{<e_c}$. During training, the entire audio condition is randomly dropped with a fixed probability (we use $10\%$) to promote robust lip-sync. 

\subsection{Long-Range Motion Prediction via Diffusion Forcing}
\label{sec:df}
We identify two key challenges in our autoregressive diffusion training setup (Equation~\ref{eq:tf}) for long-sequence motion generation. First, prediction errors accumulate over time: early inaccuracies propagate through the sequence, degrading subsequent facial expressions and potentially leading the model to internalize spurious patterns, resulting in unstable long-term synthesis. This arises from the discrepancy between training and inference—during training, ground-truth motion history is available, whereas in inference, the model relies on its own imperfect predictions. Second, the training scheme defined in Equation~\ref{eq:tf} does not support chunk-wise parallelization, unlike typical autoregressive models, leading to inefficient training.

To address these issues, we introduce a diffusion-forcing mechanism~\cite{chen2025diffusion,song2025historyguidedvideodiffusion}. Specifically, the entire motion sequence $\mathbf{z}^{1:T}$ is corrupted using independently sampled noise timesteps $k^c$ across chunks, providing noisy historical motion context even during training. This reduces the model’s over-reliance on exact ground-truth history, mitigating error accumulation and encouraging more robust, diverse, and adaptive long-range motion predictions.  Crucially, this mechanism also shifts the model’s attention more strongly toward audio cues, preventing overfitting to motion priors and improving both expressive dynamics and lip synchronization.
Importantly, noise levels are sampled independently across all chunks, without distinguishing between past and future segments. This uniform treatment enables parallel training of all motion chunks using the diffusion loss, similar to standard autoregressive training:
\begin{equation}
\label{eq:df}
\mathbb{E}_{\mathbf{z}^{1:T}, \epsilon_{k^C}, k^C} \left[\frac{1}{C} \sum_{c=1}^{C} \left\|\,\mathbf{v}_{k^c} - v_\theta\left(\mathbf{z}_{k^c}^{s_c:e_c}, k^c,  \{\mathbf{z}_{k^i}^{s_i:e_i}\}^{i<c}, \mathbf{a}^{<e_c}\right)\,\right\|^2\right].
\end{equation}

\noindent\textbf{Inter-Chunk Asynchronous Diffusion. } Within each chunk, we adopt an asynchronous diffusion scheme~\cite{chen2024asyncdiff} where each latent is assigned distinct noise level, rather than sharing a uniform timestep. This integrates seamlessly with our diffusion-forcing scheme and enables fine-grained control over history context guidance at inference. Empirically, it leads to richer dynamics, more fluid emotional transitions, and superior long-form temporal coherence. 
\begin{figure*}[ht]
    \centering
    \includegraphics[width=0.82 \textwidth]{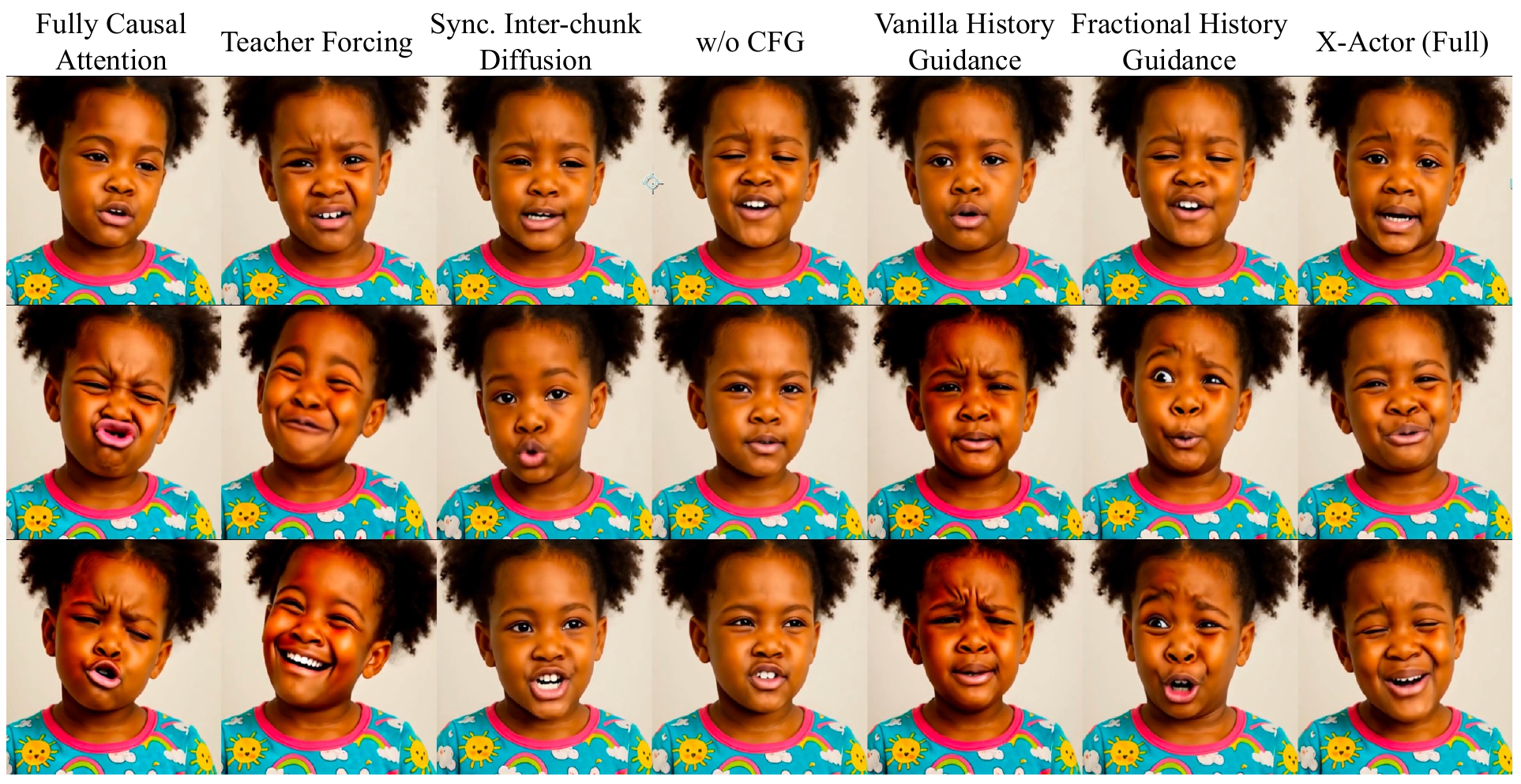} 
    \vspace{-3mm}
    \caption{\textbf{Qualitative ablation.} Each column represents a generated sequence of each ablated model, driven by the same audio.}
    \vspace{-3mm}
    \label{fig:abla}
\end{figure*}

\subsubsection{Inference Strategies}
At inference time, we adopt the DDIM scheduler~\cite{song2021denoising} to denoise each motion chunk, employing Classifier-Free Guidance (CFG) by performing two forward passes—one with audio conditioning and one without. 
To incorporate long-range motion context, our diffusion-forcing training scheme allows for flexible scheduling of historical noise levels, enabling a tunable balance between motion dynamics, temporal stability, and output diversity. Specifically, we apply a monotonically decreasing noise schedule over historical motion context: the noise timestep $k$ is linearly decreased from a maximum value (i.e., Gaussian noise) for distant frames down to 0 for frames closest to the current one.
This strategy ensures that nearby frames contribute clean and fine-grained motion cues, fostering smooth transitions and local coherence, while distant frames provide coarser, low-frequency signals that promote global emotional consistency and continuity.

%% file: sections/experiments.tex
\vspace{-1mm}
\section{Experiments}
\begin{table*}[t!]
\centering
\caption{Comparisons with SOTA methods on the self/cross-driven testsets. We highlight the \textbf{best} and \underline{second-best} scores using bold and underline, respectively.}
\vspace{-3mm}
\label{tab:quant_reenact}
\scalebox{0.95}{
\begin{tabular}{lcccccc|cccc|cccc}
\toprule
\multirow{2}{*}{\textbf{Methods}} 
& \multicolumn{6}{c|}{\textbf{Self-Driven (RAVDESS~\cite{RAVDESS})}} 
& \multicolumn{8}{c}{\textbf{Cross-Driven (In-the-Wild)}} \\
\cmidrule(lr){2-7} \cmidrule(lr){8-15}

& \textbf{SynC}$\uparrow$ 
& \textbf{SynD}$\downarrow$
& \textbf{FVD}$\downarrow$ 
& \textbf{DGlo}$\downarrow$
& \textbf{DExp}$\downarrow$ 
& \textbf{DEmo}$\downarrow$ 
& \textbf{SynC}$\uparrow$ 
& \textbf{SynD}$\downarrow$ 
& \textbf{Glo}$\uparrow$ 
& \textbf{Exp}$\uparrow$ 
& \textbf{Syn}$\uparrow$ 
& \textbf{Div}$\uparrow$ 
& \textbf{Emo}$\uparrow$ 
& \textbf{VQ}$\uparrow$ 
\\
\midrule

SadTalker     & 5.26 & 7.28 & 463.1 & 0.011 & 0.0016 & 0.49 & 5.17 & 8.09 & 0.064 & 0.015 & 0.015 & 0.029 & 0.025 & 0.025 \\ 
JoyVasa      & 4.55 & 9.19 & 359.2 & 0.084 &	0.0034 & 0.48 & 5.18 & 8.87 & 0.002 & 0.007  & 0.000 & 0.000 & 0.005 & 0.000 \\
EchoMimic    & 4.39 & 8.56 & 674.8 & 0.028 & 0.0069 & 0.45 & 3.79 & 9.86 & 0.035 & 0.018  & 0.060 & 0.059 & 0.060 & 0.060 \\
Hallo3 & 4.84 & 8.40 & 290.1 & 0.039 & 0.011 & 0.47 & 4.48 & 9.74 & 0.138 & 0.016  & 0.015 & 0.015 & 0.015 & 0.015 \\
MEMO      & 5.53 & 8.12 & 279.9 & 0.076 & 0.003 & 0.44 & 5.50 & 8.6 & 0.022 & 0.01  & 0.035 & 0.044 & 0.045 & 0.045 \\
Sonic         & 5.99 & 7.43 & 230.8 & 0.04 & 0.0019 & 0.46 & 6.28 & 7.64 & 0.032 & 0.009  & 0.114 & 0.112 & 0.130 & 0.119 \\
\midrule
\textbf{X-Actor} 
              & \textbf{6.33} & \textbf{7.21} & \underline{278.3} & \underline{0.021} & 0.005 & \textbf{0.37} & \underline{5.63} & 8.72 & \textbf{0.144} & \textbf{0.034}  & \textbf{0.761} & \textbf{0.741} & \textbf{0.720} & \textbf{0.736} \\
\bottomrule
\end{tabular}}
\vspace{-3mm}
\label{tab:comp}
\end{table*}

\subsection{Implementation Details}
\noindent\textbf{Datasets.} We curate a large-scale training corpus by collecting talking head videos from multiple public datasets—HDTF~\cite{zhang2021flow}, CelebV-HQ~\cite{zhu2022celebvhq}—and additional sources from online platforms. To ensure high-quality audio-visual alignment, we filter out clips with low lip-sync scores using SyncNet~\cite{chung2017out}. The final dataset comprises 160K short clips (each under 10 seconds, averaging 5 seconds), totaling 260 hours, and 280K long-form clips (ranging from 10 seconds to 1 minute, averaging 15 seconds), totaling 900 hours of talking video data.

For evaluation, we consider two settings. In the self-driven setting, we randomly sample a frame from RAVDESS~\cite{RAVDESS} video clips and animate it using a different audio clip from the same actor, resulting in 100 identity-consistent audio-visual pairs. To assess performance in more challenging and long-form conditions, we additionally construct an in-the-wild setting, comprising 50 diverse portrait images—spanning various identities, styles, expressions, and head poses—and 50 emotionally rich multilingual audio clips (20 to 200 seconds)  from movies, TV shows, and songs.

\noindent\textbf{Training and Inference.}
We train our framework in two stages. In the first stage, following X-NeMo~\cite{xnemo2025}, we jointly train the latent motion representation and a video diffusion model composed of the SD1.5 UNet~\cite{rombach2022high}, ReferenceNet~\cite{hu2024animate,cao2023masactrl}, and temporal modules~\cite{guo2023animatediff}. This stage is conducted on 8 A100 GPUs for 120K steps with a batch size of 8. In the second stage, we fine-tune the 7B Qwen backbone~\cite{qwen2.5} augmented with newly inserted audio cross-attention layers. This stage is trained on 16 A100 GPUs for 66K steps using a learning rate of $1e^{-4}$ and a runtime-dynamic batch size to accommodate variable-length sequences.
We set the maximum temporal context length to 512 frames, though our model is scalable to longer contexts given sufficient high-quality long-form talking data. The chunk size for motion generation is set to 64. During inference, we use a CFG scale of 3 and apply a 50-step DDIM scheduler~\cite{song2021denoising} to denoise the motion latents. The final video is synthesized from the predicted motion latents and the reference image using a 25-step DDIM scheduler with a CFG scale of 2.5.

\vspace{-2mm}
\subsection{Evaluations and Comparisons}
We evaluate our method against state-of-the-art audio-driven portrait animation baselines, including both end-to-end diffusion models—such as Hallo3~\cite{cui2024hallo3}, MEMO~\cite{zheng2024memo}, and Sonic~\cite{ji2024sonic}—and staged generation methods. Among the staged methods, EchoMimic~\cite{chen2025echomimic} predicts audio-driven facial landmarks that control a subsequent diffusion model. JoyVasa~\cite{cao2024joyvasaportraitanimalimage}, an open-source implementation of VASA-1~\cite{xu2024vasa}, and SadTalker~\cite{zhang2023sadtalker} employs implicit facial motion latents and synthesize videos using GAN-based renderers.
We exclude FADA~\cite{zhong2025fada} 
from quantitative comparisons due to the lack of publicly available code and models, but we include some qualitative comparison. All evaluation metrics are computed on generated videos at a resolution of $512 \times 512$.

\begin{table}
\caption{Quantitative ablation study.}
\vspace{-3mm}
    \centering
    \begin{tabular}{l|cc}
    \hline
    Method                  &  \textbf{SynC$\uparrow$} &  \textbf{SynD$\downarrow$} \\ \hline
     fully causal attention                      & 4.04 & 9.39    \\
     teacher-forcing scheme        & 5.79 & 7.67  \\
     synchronous inter-chunk diffusion     & 6.30 & 7.51   \\
     w/o CFG               & 3.72 & 9.65  \\
     vanilla history guidance                & 1.65 & 11.34   \\
     fractional history guidance                & 2.60 & 11,19   \\    
    \midrule
     \textbf{\papername}                            & \textbf{6.62} & \textbf{7.18}  \\ \hline
    \end{tabular}
\vspace{-5mm}
\label{tab:abla}
\end{table} 

\noindent\textbf{Quantitative Evaluation.} We use six metrics for quantitative evaluation, with ground-truth comparisons available in the self-driven setting. SynC$\uparrow$ and SynD$\downarrow$\cite{chung2017out} are used to assess the synchronization between the input audio and lip movements. Fréchet Video Distance (FVD$\downarrow$)\cite{unterthiner2019fvd} measures the overall visual quality of generated videos.
To evaluate expressiveness beyond lip sync, we use Global Motion (Glo$\uparrow$) and Dynamic Expression (Exp$\uparrow$) metrics~\cite{jiang2024loopy}, which quantify motion variance of keypoints in the nose and upper face regions—excluding the mouth—to capture naturalistic head and facial dynamics. For self-driven evaluations, we report DGlo$\downarrow$ and DExp$\downarrow$, the L1 differences between generated and ground-truth motion statistics for global movement and expressions, respectively.
Finally, we use EMONet~\cite{toisoul2021estimation} to predict perceived emotions and compute DEmo$\uparrow$, the L1 difference between predicted and ground-truth emotion vectors, to assess emotional alignment with the audio. Table~\ref{tab:comp} reports the quantitative results on both test sets. Our method outperforms all baselines on most metrics, including lip sync accuracy (SynC$\uparrow$), emotion alignment (DEmo$\downarrow$), and motion expressiveness (Glo$\uparrow$, Exp$\uparrow$).

\noindent\textbf{User Study.} We conducted a user study on our in-the-wild test set (i.e., the cross-driven setting) for subjective evaluation against competing methods across four key dimensions: lip sync (Syn$\uparrow$), motion diversity(Div$\uparrow$), emotion alignment(Emo$\uparrow$), and video quality (VQ$\uparrow$). A total of 21 participants were asked to select the best video for each dimension from a randomized set of model outputs across 10 examples. As shown in Table~\ref{tab:comp}, our method consistently outperforms all baselines across all dimensions, 
underscoring the efficacy of our autoregressive latent motion diffusion framework.

\noindent\textbf{Qualitative Evaluation.} 
Figure~\ref{fig:comp} presents a visual comparison using long-form audio inputs. Prior methods struggle with expressive motion and often fail to capture nuanced emotional transitions, particularly when the reference image displays emotions that conflict with the audio. In contrast, \papername~effectively conveys fluid, evolving emotions, producing expressive and nuanced facial motions while preserving identity and maintaining accurate lip sync across extended sequences, with minimal quality degradation. Additional dynamic comparisons can be found in the supplementary material.

\vspace{-2mm}
\subsection{Ablation Study}

We ablate the key components of our framework by systematically replacing them with alternative designs, and evaluate all variants on our in-the-wild test set, where audio-image pairs are randomly matched. Table~\ref{tab:abla} reports quantitative lip-sync scores across ablated models and our full pipeline, complemented by visual comparisons in Figure~\ref{fig:abla} and on our supplementary webpage.
First, we assess the role of our attention mechanism by replacing the full self-attention within each chunk with fully causal attention. This ablated model exhibits noticeable jittering artifacts and erratic facial motion during the initial phase due to insufficient audio context.
Next, we evaluate the impact of the diffusion-forcing strategy by substituting the autoregressive diffusion training with a teacher-forcing scheme that uses ground-truth history. The resulting model suffers from severe error accumulation at inference time, leading to degraded lip sync and non-recoverable facial expressions as the audio progresses.
In contrast, our full pipeline maintains stable, coherent long-form performance with synchronized lip movements and expressive emotional alignment.
We also examine the benefit of asynchronous noise timesteps within each chunk by comparing against standard diffusion with synchronous noise. Our inter-chunk asynchronous diffusion enables more flexible and generalizable training, resulting in improved motion dynamics with on-par lip-sync score.
Additionally, removing the audio CFG results in a notable decline in lip-sync performance, underscoring its importance in aligning facial motion with audio content.
Finally, we evaluate our time-adaptive historical context scheduler—which applies stronger noise to more distant motion history—against two baselines~\cite{song2025historyguidedvideodiffusion}: a vanilla strategy (equal guidance across all history), and the fractional scheduler(staircase-increasing guidance). As further evidenced in Figure~\ref{fig:abla}, our scheduler consistently enhances long-range emotional coherence and expressiveness in facial motion.

%% file: sections/conclusion.tex
\vspace{-2mm}
\section{Conclusion}

We introduced \papername, a novel framework for generating long-range, emotionally expressive portrait animations from audio and a single image. By decoupling motion generation and video synthesis, and integrating with a diffusion-forcing training paradigm, our method enables stable, actor-quality portrait animations with fine-grained emotion and temporal coherence. Extensive experiments demonstrate state-of-the-art performance, setting a new benchmark for audio-driven portrait actings.

\noindent\textbf{Limitations and Future Work.} 
Our method is currently limited to talking heads and does not model full-body movements or gestures. Extending to co-speech body and gesture generation is a natural next step toward holistic, actor-level performance synthesis. The expressiveness of generated videos is also constrained by current video diffusion models; future work includes enabling richer dynamics such as crying or trembling, and generalizing to non-human subjects like animals. Lastly, the scarcity of high-quality, long-form emotional acting data limits performance; we plan to explore scalable data collection and learning from in-the-wild performances.